\documentclass{article}

\usepackage{spconf}
\usepackage{graphicx,epstopdf,hyperref,subcaption}
\usepackage{float}
\usepackage{enumitem}
\usepackage{algorithm,algorithmicx,algpseudocode}
\usepackage{multirow}
\usepackage{amsthm,amsmath,amssymb,braket}
\usepackage[font=footnotesize,skip=5pt]{caption}
\usepackage{color}
\usepackage{arydshln}
\algdef{SE}[DOWHILE]{Do}{doWhile}{\algorithmicdo}[1]{\algorithmicwhile\ #1}%

\usepackage{bm}

\long\def\comment#1{}

\newfont{\bbb}{msbm10 scaled 700}

\newcommand{\fv}{{\bf f}}

\newcommand{\xv}{{\bf x}}

\newcommand{\Xc}{{\cal X}}

\newtheorem{theorem}{Theorem}

\newtheorem{lemma}{Lemma}

\title{Active Learning on Weighted Graphs\\ Using Adaptive and Non-adaptive Approaches}

\name{Eyal En Gad, Akshay Gadde, A. Salman Avestimehr and Antonio Ortega
\thanks{This work is supported in part by NSF under grants CCF-1410009 and CCF-1527874}\vspace{-2mm}
}

\address{Department of Electrical Engineering\\
	University of Southern California, Los Angeles\vspace{-2mm}}
    
\begin{document}
\ninept

\maketitle
\begin{abstract}
This paper studies graph-based active learning, where the goal is to reconstruct a binary signal defined on the nodes of a weighted graph, by sampling it on a small subset of the nodes. 
A new sampling algorithm is proposed, which sequentially selects the graph nodes to be sampled, based on an aggressive search for the boundary of the signal over the graph. The algorithm generalizes a recent method for sampling nodes in unweighted graphs. The generalization improves the sampling performance using the information gained from the available graph weights. 
An analysis of the number of samples required by the proposed algorithm is provided, and the gain over the unweighted method is further demonstrated in simulations. Additionally, the proposed method is compared with an alternative state-of-the-art method, which is based on the graph's spectral properties. It is shown that the proposed method significantly outperforms the spectral sampling method, if the signal needs to be predicted with high accuracy. On the other hand, if a higher level of inaccuracy is tolerable, then the spectral method outperforms the proposed aggressive search method. Consequently, we propose a hybrid method, which is shown to combine the advantages of both approaches.
\end{abstract}
\begin{keywords}
active learning on graphs, adaptive and non-adaptive sampling of graph signals, sampling complexity
\end{keywords}

\section{introduction}
This paper studies the problem of binary label prediction on a graph. In this problem, we are given a graph $G = (V,E)$, where the edges $E$ (which can be weighted) capture the similarity relationship between the objects represented by the nodes $V$. Each node has an initially unknown label associated with it, given by a signal $f:V \rightarrow \{-1,+1\}$. 
%
The goal is to reconstruct the entire signal by sampling its values on a small subset of the nodes. This is achievable when the signal bears some degree of smoothness over the graph, which means that similar objects are more likely to have the same label. 
%
Active learning aims to minimize the number of samples needed by selecting the most informative nodes.

This problem arises in many machine learning applications, where there is an abundance of unlabeled data but labeled data is scarce and expensive to obtain, for example, requiring human expertise or elaborate experiments. Active learning is 
an  effective way to minimize the cost of labeling in such scenarios~\cite{Settles-UWM-10}. A graph based approach to this problem starts by creating a graph where the nodes correspond to the data points $\Xc = \{\xv_1, \ldots, \xv_n\}$ and the edges capture the similarity between them. Typically, a sparse graph 
which connects each data point to few of its most similar neighbors is used. 
%
%
The unknown labels $f_i \in \{-1,+1\}$ associated with the data points define a binary function on the nodes. 
In datasets of interest, the signal is often notably smooth on the graph. 

There are two approaches to active learning on graphs. The first approach focuses on \emph{identifying the nodes near the boundary} region, where the signal changes from $+1$ to $-1$. The methods with this approach~\cite{DasNowZhu15,Zhu-ACTIVE-ICML-03} sample nodes sequentially, i.e., the nodes to be sampled next are chosen based on the graph structure as well as previously observed signal values. 
The second approach~\cite{Anis-ICASSP-14,Ji-AIST-12,Guillory-UAI-11}, in contrast, utilizes \emph{global} properties of the graph in order to identify the most informative nodes, and sample them all at once. Such global approaches usually focus on providing a good \emph{approximation} of the signal, rather than exact recovery. It is also possible to combine the two approaches, for example, as in~\cite{Osugi-ICDM-05}.


The first contribution of this paper is a new sampling algorithm, called weighted $S^2$ which takes the boundary sampling approach. The weighted $S^2$ algorithm is a generalization of a recently proposed algorithm called $S^2$~\cite{DasNowZhu15}, which 
is defined only for the case of unweighted edges. The purpose of the generalized algorithm is to take advantage of the additional information available in the form of the edge weights, in order to reduce the sampling complexity. We characterize the sampling budget required for signal recovery by the weighted $S^2$ algorithm, as a function of the complexity of the signal with respect to the graph. We explain how this generalization can be useful in reducing sampling complexity, and demonstrate a significant reduction (nearly $25\%$ in one dataset), when neighboring nodes of opposite labels are considerably less similar to each other than identically labeled neighbors.





We further compare the sampling complexity of the weighted $S^2$ algorithm with an alternative state-of-the-art method called the cutoff maximization method~\cite{Anis-ICASSP-14}. 
Unlike the $S^2$ methods, which aim for a complete recovery of the signal by aggressively searching for the boundary nodes, the cutoff maximization method is focused only on providing a good \emph{approximation} of the signal by ensuring that the unsampled nodes are well-connected to the sampled nodes~\cite{Gadde-KDD-14}.
This method 
finds a sampling set by optimizing a spectral function defined using the graph Laplacian.  
We perform the comparison on three realistic data sets, and observe two interesting results: 

1.~The cutoff maximization method does not discover the entire boundary (i.e, nodes with oppositely labeled neighbors) unless the sampling set contains almost all the nodes. In contrast, the number of nodes required by the $S^2$ methods to discover the boundary is not considerably larger than the number of boundary nodes.

2.~When the sampling budget is quite limited, the cutoff maximization method provides a much better approximation of the signal. 
There exists a threshold in terms of the sampling budget
that determines which method offers better accuracy. Conversely, the tolerable degree of inaccuracy determines which of the methods offer lower sampling complexity.
%

Motivated by the second observation, we propose a hybrid approach (similar in spirit to \cite{Osugi-ICDM-05}) which samples the first few nodes with the cutoff maximization method to approximate the boundary and then switches to the weighted $S^2$ method to refine the approximation. The experiments suggest that the hybrid approach combines the advantages of both methods.

\vspace{-0.5\baselineskip}
\section{$S^2$ Algorithm for Weighted Graphs}
\label{sec:algo}
\vspace{-0.25\baselineskip}
The $S^2$ algorithm was proposed in~\cite{DasNowZhu15} for unweighted graphs. In this section we describe the principle of the algorithm, and then generalize the algorithm to weighted graphs. In the next section we analyze the query complexity of the generalized algorithm. 

\vspace{-0.75\baselineskip}
\subsection{Original $S^2$ Algorithm}
The goal of the algorithm is to find the signal $f$. To do this, the algorithm operates by finding the edges that connect oppositely labeled nodes. These edges are called cut edges, and together the set of cut edges is called the cut. The algorithm incrementally identifies the cut, with the rationale that once the entire cut is identified, the signal is completely recovered. To find the cut, the algorithm maintains a copy of the graph $G$, and each time it samples a node that neighbors previously sampled nodes of opposite label, it removes the newly discovered cut edges from the graph copy. This way, the remaining graph copy contains only the undiscovered part of the cut, and the algorithm can more easily focus on discovering these edges.

The main idea of the algorithm is to look for a pair of oppositely labeled nodes, and then to find the cut edge on the shortest path between the nodes, using a binary search procedure. The algorithm begins with a random sampling phase. At this phase the algorithm queries a random node, according to the uniform distribution. After each sample, the algorithm checks whether the sampled node has any previously sampled neighbors of opposite label. If such neighbors exist, then the connecting edges are newly discovered cut edges, and they are removed from the graph. After checking and potentially removing newly discovered cut edges, the algorithm checks whether the remaining graph contains any pair of connected nodes of opposite labels. If no such pair exists, the algorithm proceeds with a random sample. If pairs of connected, oppositely labeled nodes do exist, the algorithm looks for the shortest path among all the paths connecting such pairs, and sample the node in the middle of that path (breaking ties arbitrarily). After each sampling operation, either a random sample or a bisecting sample, the algorithm again removes all newly discovered cut edges.

The $S^2$ algorithm is described more formally in Algorithm~\ref{alg:s2}. The algorithm is given a budget, which determines the number of queries to perform. Once the budget is exhausted, the algorithm calls a label completion function, which predicts the labels of the unsampled nodes. Several such label completion algorithms are known, such as the POCS method in \cite{Narang-GlobalSIP-13}. The $S^2$ algorithm uses a function called Middle Shortest Path (MSP), which returns the node in the middle of the shortest path among all the paths connecting a pair of oppositely labels nodes in the remaining graph.

\begin{algorithm}
\caption{$S^2$ Algorithm}
\label{alg:s2}
\begin{algorithmic}[1]
\Statex \textbf{Inputs}: Graph $G$, BUDGET $\le n$.
\State $L\leftarrow \emptyset$
\While {$1$}
\State $x\leftarrow$ Randomly chosen unlabeled node.
\Do
\State Add $(x,f(x))$ to $L$.
\State Remove newly discovered cut edges from $G$.
\If {$|L| =\text{BUDGET}$  }
\State \Return LabelCompletion$(G,L)$
\EndIf
\doWhile $x\leftarrow \text{MSP(G,L)}$ exists
\EndWhile
\end{algorithmic}
\end{algorithm}

\vspace{-0.75\baselineskip}
\subsection{Generalization for Weighted Graphs}
The $S^2$ algorithm in~\cite{DasNowZhu15} is defined only for unweighted graphs. Since many learning scenarios provide a weighted graph, we extend the algorithm to exploit the additional available information by modifying the MSP function in the algorithm. 
Our modification is based on the assumption that the signal is smooth,
which means that high-weight edges connect mostly nodes of the same label. Therefore, the weight of cut edges is generally low. 

\begin{figure}[t]
\centering
\includegraphics[width=0.42\textwidth]{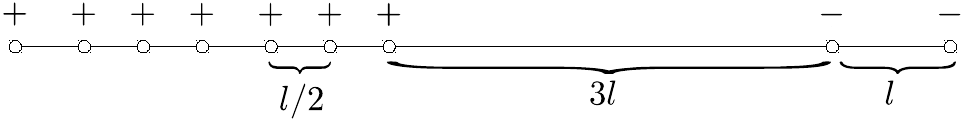}
\caption{An illustration of advantage of weighted $S^2$ over unweighted $S^2$.}
\label{fig:ws2_adv}
\vspace{-\baselineskip}
\end{figure}

In the unweighted $S^2$ algorithm, each MSP query reduces the number of nodes in the shortest of the paths between any two oppositely labeled nodes by approximately one half. 
The main idea in our generalization is to take advantage of the low weights of cut edges in order to reduce this number 
by more than a half with each query. To do this, we first switch our perspective, for convenience, from the edge weights to the distances associated with the edges, which are inversely proportional to the weights. The distance between non-neighboring nodes is defined as the sum of the lengths of the edges in the shortest path connecting the nodes.
Since the weights are a decreasing function of the distances, it follows that \emph{cut edges are typically longer than other edges}. We take advantage of this fact by modifying the MSP function to \textbf{sample the node 
closest to the midpoint of the path, where the midpoint is computed in terms of length, rather than the number of edges in the path}. 
%
With each query, the proposed sampling rule can potentially reduce the number of nodes along the shortest of the paths between any two oppositely labeled nodes by more than half if the cut edges contribute significantly more than the non-cut edges to the length of the path. 
Thus, ultimately it requires less samples to discover a cut edge. 
This intuition is demonstrated in Figure~\ref{fig:ws2_adv}. In this example, the nodes labeled $+1$ are connected with an edge of length $l/2$, the nodes labeled $-1$ are connected with an edge of length $l$ and the cut edge is of length $3l$.
%
%
Given the labels of the end nodes of this path, the binary search phase of the unweighted $S^2$ algorithm needs to sample labels of $3$ extra nodes to discover the cut edge. The weighted $S^2$ algorithm, on the other hand, finds the cut edge with only $2$ samples. 
This type of situation arises more prominently in an unbalanced data set, where the number of nodes in one class is much larger than the other. 
The advantage of weighted $S^2$ algorithm in such a case is experimentally verified in Section~\ref{sec:experiments}.







\vspace{-0.5\baselineskip}
\section{Analysis of the Weighted $S^2$ Algorithm}
\vspace{-0.5\baselineskip}
\subsection{Notation}
Note that $f$ partitions the vertices of $G$ into a collection of connected components with identically labeled vertices.
Let $V_1,V_2,\dots,V_k$ be these $k$ connected components.
Notice that the first node that $S^2$ queries in each collection $V_i$ is often queried randomly, and not by a bisection query.
Define 
\vspace{-0.95\baselineskip}
$$\beta\triangleq\min_{1\le i\le k}\frac{|V_i|}{n}.$$ 
If $\beta$ is small, more random queries are required by $S^2$. 
Let $C$ be the set of cut edges in $G$. The length of the shortest cut edge in $G$ is denoted by $l_{\text{cut}}$. Let $\partial C$ be the set of nodes which share an edge with at least one oppositely labeled node. The nodes in $\partial C$ are called boundary nodes.

For $1\le i<j\le k$, let $C_{i,j}$ be the subset of $C$ for which each edge $\{x,y\}\in C_{i,j}$ satisfies $x\in V_i$ and $y\in V_j$. If $C_{i,j}$ is not empty for some $1\le i<j\le k$, then it is called a cut component. The number of cut components is denoted by $m$.
For a pair of nodes $v_1$ and $v_2$, let $d(v_1,v_2)$ denote the length of the shortest path between $v_1$ and $v_2$.
Define $l_n=\max_{v_1,v_2\in V}d(v_1,v_2)$. 

\vspace{-0.5\baselineskip}
\subsection{Cut Clustering}
For nodes $x,y\in V$, let $d^G(x,y)$ be the length of the shortest path connecting $x$ and $y$ in $G$.
Let $e_1=\{x_1,y_1\}$ and $e_2=\{x_2,y_2\}$ be a pair of cut edges in $G$ such that $f(x_1)=f(x_2)$ and $f(y_1)=f(y_2)$. 
Define 
$$\delta(e_1,e_2)=d^{G-C}(x_1,x_2)+d^{G-C}(y_1,y_2)+\max\{l_{e_1},l_{e_2}\},$$
 where $G-C$ is the graph $G$ with all the cut edges removed. Let $H_r=(C,\mathcal{E})$ be the meta graph whose nodes are the cut edges of $G$, and $\{e,e'\}\in\mathcal{E}$ iff $\delta(e,e')\le r$. Let $l_{\kappa}$ be the smallest number for which $H_{l_{\kappa}}$ has $m$ connected components. The motivation for the definition of $l_{\kappa}$ is demonstrated in the following lemma.


\begin{lemma}
\label{lem:kappa}
Consider a case in which after the removal of an edge $e$ by the weighted $S^2$ algorithm, there exist an undiscovered cut edge $e'$ in the same cut component of $e$. Then the length of the shortest path 
between two oppositely labeled nodes 
in the remaining graph is at most $l_{\kappa}$.
\end{lemma}
\begin{proof}

Consider the connected component of the meta graph $H_{l_{\kappa}}$ that contains the removed cut edge (as a meta node). 
By the assumptions of the lemma, there must be at least one meta edge in this connected component that connects a discovered cut edge to an undiscovered cut edge. This meta edge corresponds to a path length at most $l_{\kappa}$ in the remaining graph, proving the lemma.
\end{proof}

\subsection{Query Complexity}


\begin{theorem}
\label{th:comp}
Suppose that a graph $G=(V,E)$ and a signal $f$ are such that the induced cut set $C$ has $m$ components with cut clustering $l_{\kappa}$. Then for any $\epsilon > 0$, the weighted $S^2$ will recover $C$ with probability at least $1-\epsilon$ if the \emph{BUDGET} is at least 
\begin{equation*}
\small
m\left\lceil 2\log_2 \left(\frac{l_n}{l_{\text{cut}}}\right)\right\rceil 
+(|\partial C|-m)\left\lceil 2\log_2 \left(\frac{l_{\kappa}}{l_{\text{cut}}}\right)\right\rceil
+\frac{\log(1/(\beta\epsilon))}{\log(1/(1-\beta))}.
\end{equation*}
\end{theorem}

The proof of Theorem~\ref{th:comp} uses the fact that after a pair of connected and oppositely labeled nodes is found, the number of queries until a boundary node is sampled is at most logarithmic in $(l/l_{\text{cut}})$, where $l$ is the length of the path between the nodes. We show this fact in the following lemma.

\begin{lemma}
\label{lem:binary}
The cut-edge of length $l_\text{cut}$ is found after no more than $r = \left\lceil{2\log_2\left(\frac{l}{l_\text{cut}}\right)}\right\rceil$ aggressive steps. 
\end{lemma}
\begin{proof}
%
\begin{figure}[t]
\centering
\includegraphics[width = 0.27\textwidth]{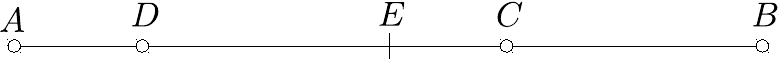}
\caption{The interval of interest is at least halved after two queries.}
\label{fig:path_bisection}
\vspace{-\baselineskip}
\end{figure}
%
%
In Figure~\ref{fig:path_bisection}, let $C$ and $D$ denote the queried nodes, such that $C$ is sampled first. Let $E$ denote the midpoint of the interval between $A$ and $B$, where there may not be a node.
By considering all the cases based on labels of $C$ and $D$ and their positions relative to $E$, 
it can be shown that after two queries, the length of the interval of interest (i.e., the interval containing the cut-edge) is at least halved. The details are omitted in the interest of space.



After $i$ (where $i$ is even) queries, the length of the interval of interest is at most $\frac{l}{2^{i/2}}$. Note that the cut-edge is found when the length of the interval of interest is less than or equal to $l_\text{cut}$. If $r$ is the maximum number of queries required to locate the cut-edge, then
\begin{equation}
\frac{l}{2^{r/2}} = l_\text{cut} \Rightarrow r = \left\lceil{2\log_2\left(\frac{l}{l_\text{cut}}\right)}\right\rceil
\end{equation}
\end{proof}

\vspace{-1.5\baselineskip}
\begin{proof}[Proof of Theorem~\ref{th:comp}]
The random sampling phase follows the same argument as in~\cite{DasNowZhu15}, which gives the term $\frac{\log(1/(\beta\epsilon))}{\log(1/(1-\beta))}$. 

A sequence of bisection queries that commences after a random query or an edge removal, and terminates with an edge removal, is call a \emph{run}. In each run, at least one boundary node is being queried. Therefore, the number of runs is no greater than $|\partial C|$.
In each cut component, after the first cut edge and boundary node are discovered, the rest of the boundary nodes are discovered in $\left\lceil{2\log_2\left(\frac{l_{\kappa}}{l_\text{cut}}\right)}\right\rceil$ queries each, according to Lemmas~\ref{lem:kappa} and~\ref{lem:binary}.
For discovering the first cut edge in each cut components, we trivially bound the number of queries by $\left\lceil{2\log_2\left(\frac{l_{n}}{l_\text{cut}}\right)}\right\rceil$. Since there are $m$ cut components, we bound the total number of bisection queries by
\begin{equation*}
m\left\lceil 2\log_2 \left(\frac{l_n}{l_{\text{cut}}}\right)\right\rceil+(|\partial C|-m)\left\lceil 2\log_2 \left(\frac{l_{\kappa}}{l_{\text{cut}}}\right)\right\rceil.
\end{equation*}
\end{proof}

\vspace{-2\baselineskip}
\section{Experiments}
\label{sec:experiments}
\begin{table*}[t]
\centering
\caption{Number of samples needed to discover all the cut edges. $S^2$ methods involve random sampling. Average over $30$ trials is reported.}
\begin{tabular}{c|c|c|c|c||c|c|c|c:c}
\hline
Data & n & $|C|$ & $|\partial C|$ & $\frac{\text{mean}(l_\text{cut})}{\text{mean}(l_\text{non-cut})}$ & Unweighted $S^2$ & Weighted $S^2$ & Cutoff & Hybrid & $n_\text{switch}$ \\ \hline
Two circles & 1000 & 129 & 160 & 4.0533 & 237 & 179.2 & 999 & 272 & 128\\		
7 v 9	& 400 & 154	& 180 & 1.1074 & 312.37 & 312.07 & 399 & 277 & 47\\

2 v 4 & 400 & 29 & 39 & 1.1183  & 49.13 & 48.37 & 394 &	76 & 38	\\

Baseball v Hockey	& 400 & 255 & 235 & 1.0691 & 368.07 & 368.17 & 399 & 384 & 42 \\ 

\hline
\end{tabular}
\label{tab:cut_samples}
\vspace{-0.5\baselineskip}
\end{table*}
%
\begin{figure*}[t]
    \centering
    \begin{subfigure}[b]{0.3\textwidth}
        \includegraphics[width=\textwidth]{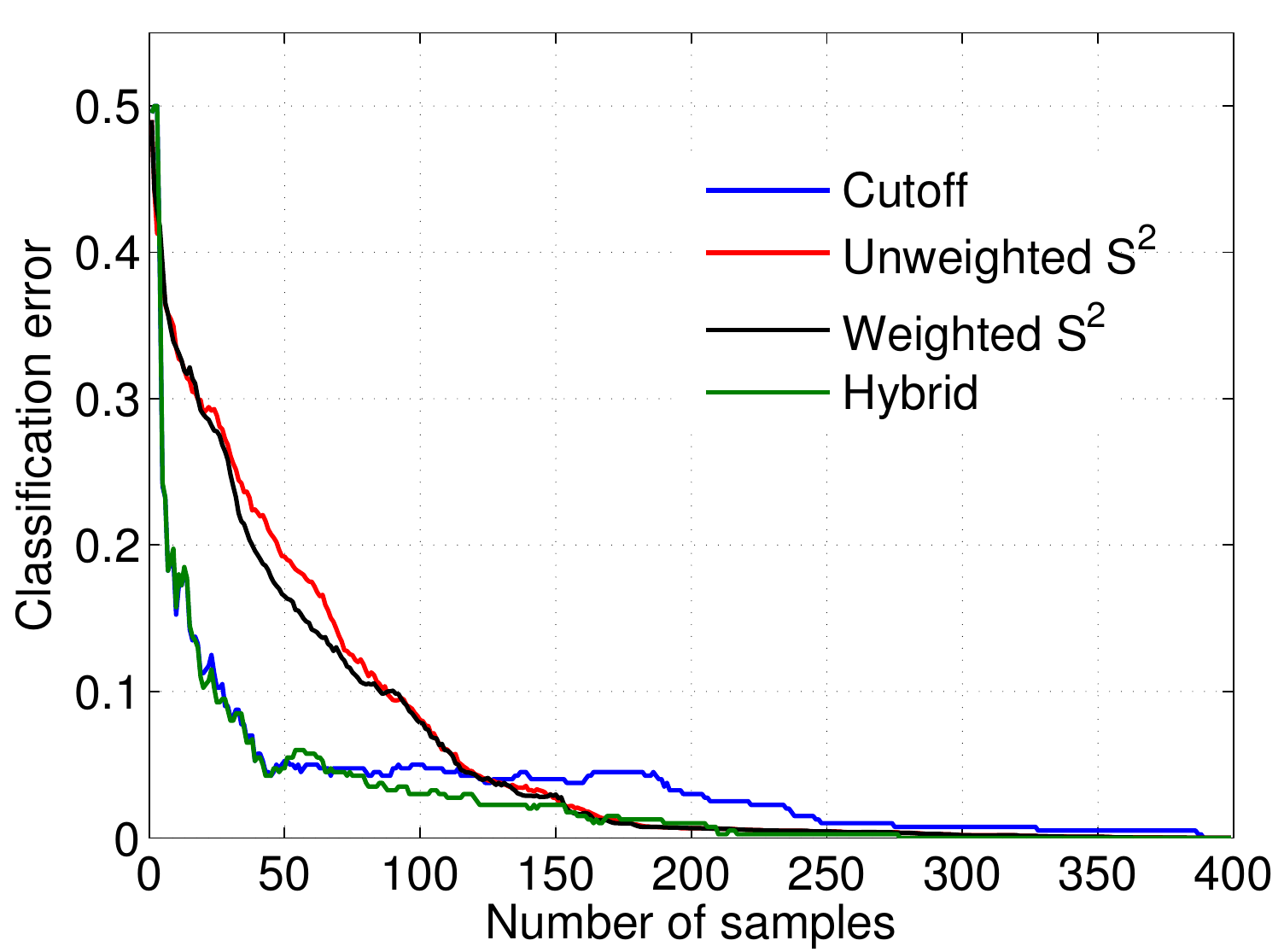}
        \caption{7 v. 9}
        \label{fig:usps}
    \end{subfigure}
    ~ 
    \begin{subfigure}[b]{0.3\textwidth}
        \includegraphics[width=\textwidth]{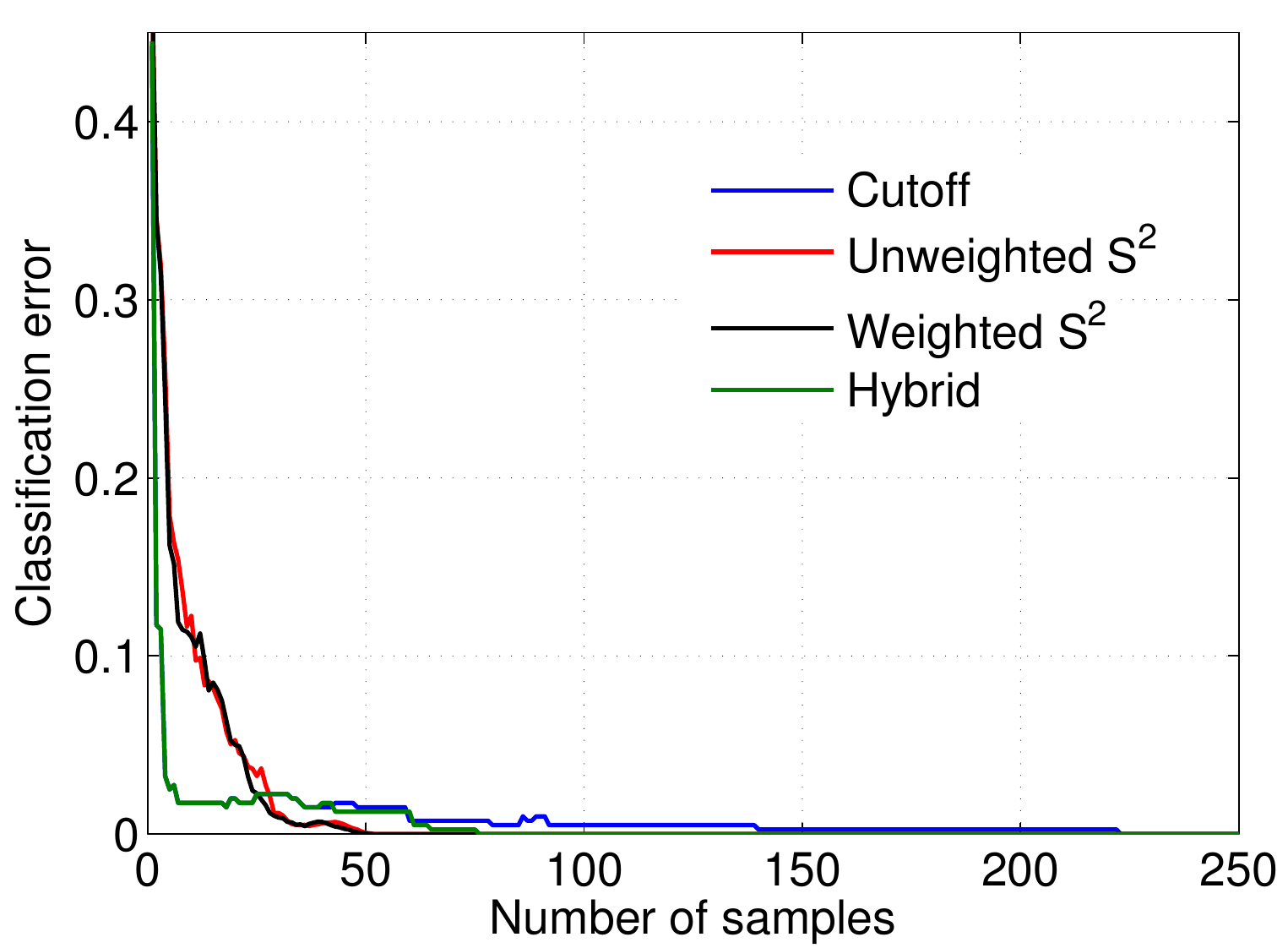}
        \caption{2 v. 4}
        \label{fig:news}
    \end{subfigure}
    ~
    \begin{subfigure}[b]{0.3\textwidth}
        \includegraphics[width=\textwidth]{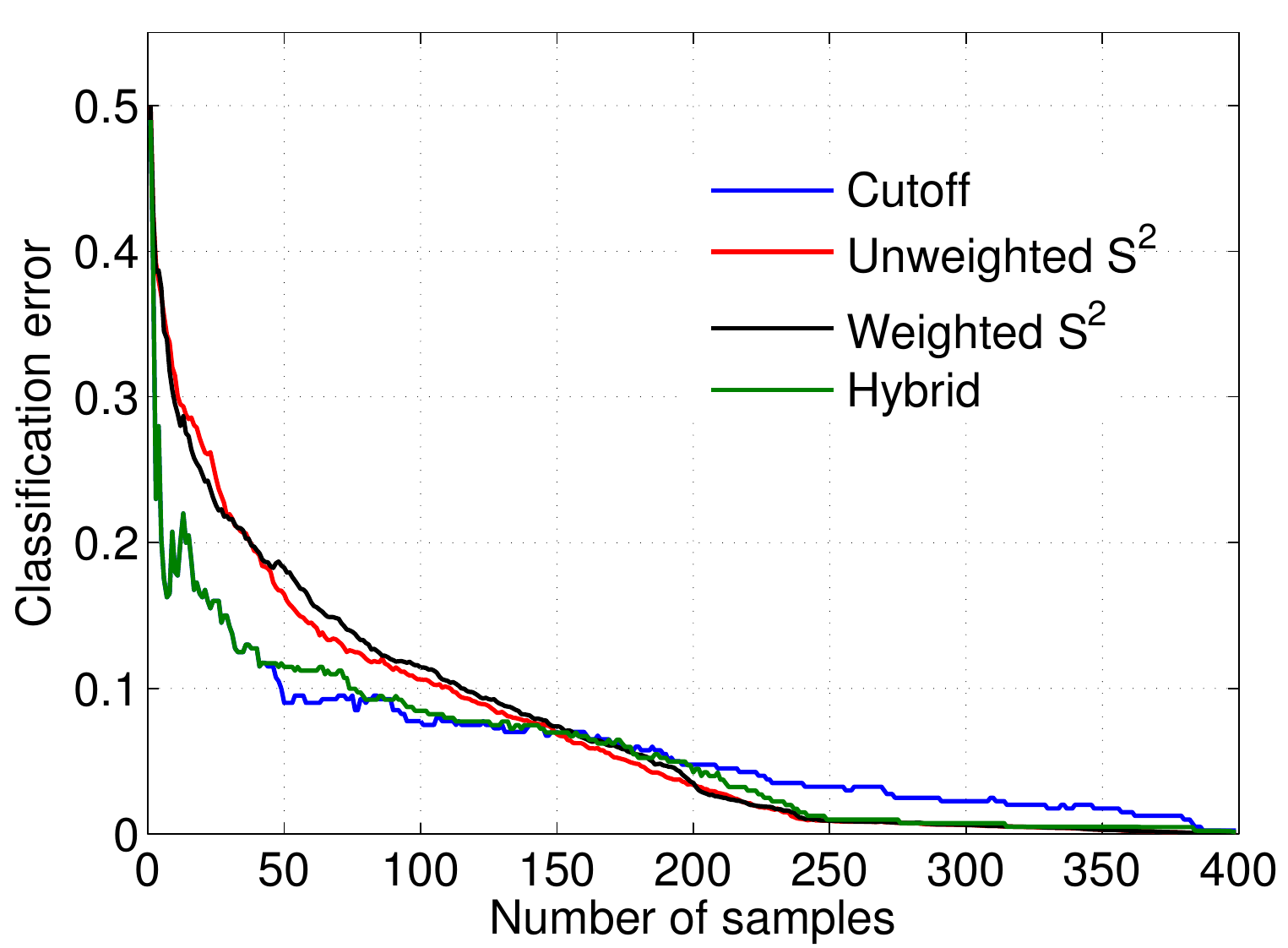}
        \caption{Baseball v. Hockey}
        \label{fig:news}
    \end{subfigure}
    \caption{Classification error against the number of sampled nodes.}
    \label{fig:error_samples}
    \vspace{-\baselineskip}
\end{figure*}
%
%
%
We consider the following graph based classification problems: 
1.~\textbf{USPS handwritten digit recognition}~\cite{USPS-dataset}: 
For our experiments, we consider two binary classification sub-problems, namely, $7$ vs. $9$ and $2$ vs. $4$, consisting of $200$ randomly selected images of each class represented as vectors of dimension $256$. 
The distance between two data points is $d(i,j) = ||\xv_i-\xv_j||$. An unweighted graph $G$ is constructed by connecting a pair of nodes $(i,j)$ if $j$ is a $k$-nearest neighbor ($k$-nn with $k = 4$)
\footnote{Due to lack of space, we do not study the effect of $k$ in detail.
} 
of $i$ or vice versa. A weighted dissimilarity graph $G_d$ is defined to have the same topology as $G$ but the weight associated with edge $(i,j)$ is set to $d(i,j)$. A weighted similarity graph $G_w$ is defined to have the same topology as $G$ but the weight associated with edge $(i,j)$ is set to $w(i,j) = \exp\left(- {d(i,j)^2}/{2\sigma^2} \right)$. 
The parameter $\sigma$ is set to be $1/3$-rd of the average distance to the $k$-th nearest neighbor for all datapoints.

2.~\textbf{Newsgroups text classification}~\cite{News-dataset}: 
For our experiments, we consider a binary classification sub-problem Baseball vs. Hockey, where each class contains $200$ randomly selected documents.
Each document $i$ is represented by a $3000$ dimensional vector $\xv_i$ whose elements are the tf-idf statistics of the $3000$ most frequent words in the dataset~\cite{Gadde-KDD-14}.
The cosine similarity between a pair data points $(i,j)$ is given by $w(i,j) = \frac{\Braket{\xv_i,\xv_j}}{||\xv_i||||\xv_j||}$. The distance between them is defined as $d(i,j) = \sqrt{1-w(i,j)^2}$. The $k$-nn unweighted graph $G$ (with $k=4$), the dissimilarity graph $G_d$ and the similarity graph $G_w$ are constructed using these distance and similarity measures as in the previous example.
%

\begin{figure}
\centering
\includegraphics[width = 0.285\textwidth]{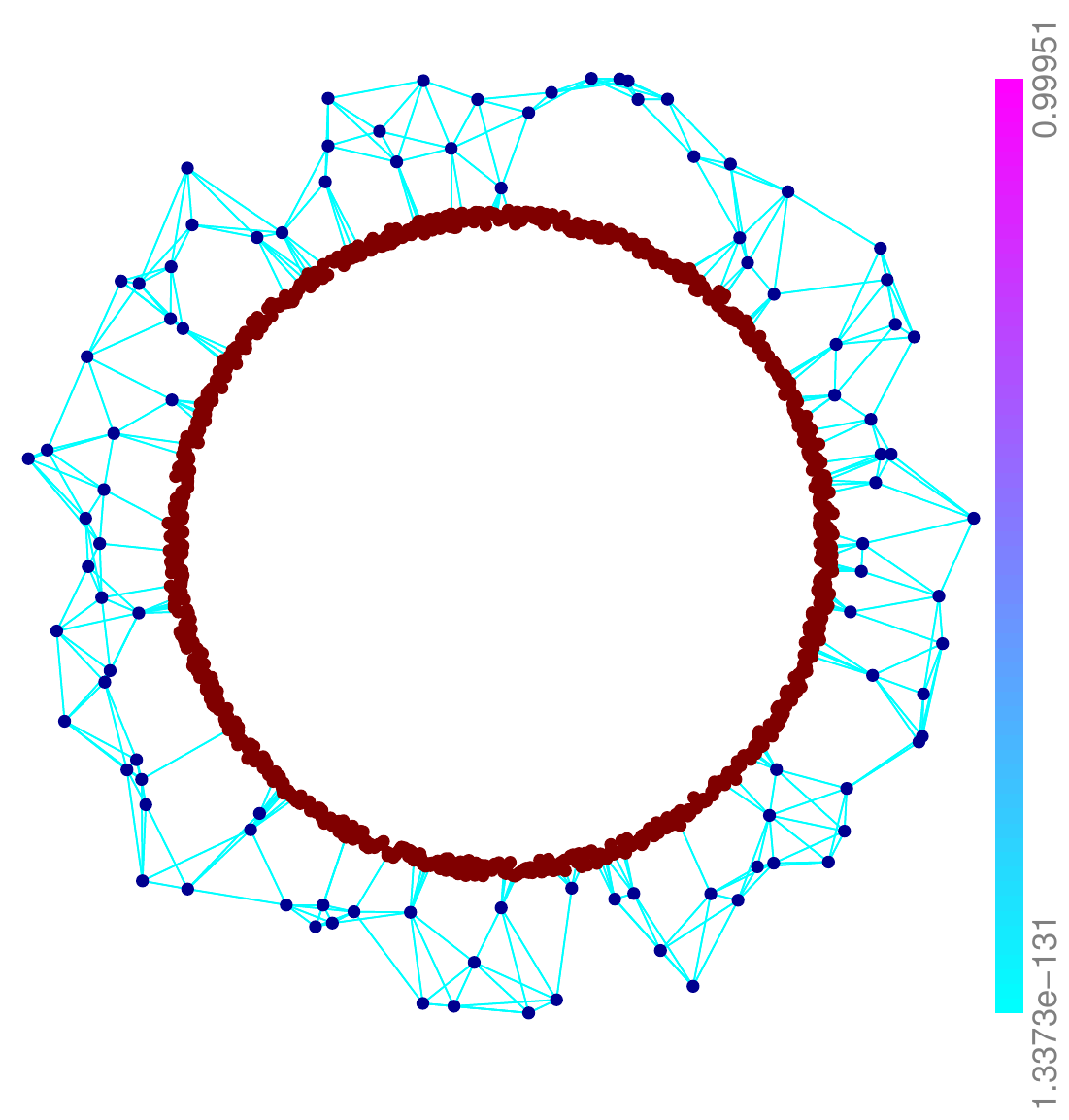}
\caption{A synthetic two circles dataset}
\label{fig:two_circles}
\vspace{-1.5\baselineskip}
\end{figure}

In addition to the above datasets, we generate a synthetic two circles dataset, shown in Figure~\ref{fig:two_circles}, in order to demonstrate the advantage of weighted $S^2$ over unweighted $S^2$. It contains $900$ points in one class (marked red) evenly distributed on the inner circle of mean radius $1$ and variance $0.05$ and $100$ points in the second class (marked blue) on the outer circle of mean radius $1.1$ and variance $0.45$. A $4$-nn graph is constructed using the Euclidean distance between the coordinates of the points.

%
We compare the performance of the following active learning methods: (1)~unweighted $S^2$ method~\cite{DasNowZhu15} with graph $G$, (2)~weighted $S^2$ method with dissimilarity graph $G_d$, (3)~cutoff maximization method~\cite{Gadde-KDD-14} with similarity graph $G_w$ and (4)~a hybrid approach combining cutoff maximization and weighted $S^2$ method. 
%
After the nodes selected by each method have been sampled,
we reconstruct the unknown label signal using the approximate POCS based bandlimited reconstruction scheme~\cite{Gadde-KDD-14} to get the soft labels. We threshold these soft labels to get the final label predictions.
The hybrid approach uses the non-adaptive cutoff maximization approach in the beginning and switches to the weighted $S^2$ method after sampling a certain number of nodes $n_\text{switch}$. In order to determine $n_\text{switch}$, after sampling the $i$-th node with the cutoff method, we compute $1-\frac{\braket{\hat{\fv}_i,\hat{\fv}_{i-1}}}{\|\hat{\fv}_{i}\|\|\hat{\fv}_{i-1}\|}$, where $\hat{\fv}_i$ denotes the vector of predicted soft labels. Once this value falls below below $0.001$, indicating that  the newly added label only marginally changed the predictions, hybrid approach switches to weighted $S^2$.
%
%

Table~\ref{tab:cut_samples} lists the number of samples required by each of the sampling methods to discover all the cut edges using the observed labels.
It shows that weighted $S^2$ can reduce the sample complexity significantly (by $25\%$) compared to unweighted $S^2$ if the ratio of mean length of cut edges and mean length of non-cut edges is high as is the case in the unbalanced two circles dataset. In rest of the datasets, the gain offered by weighted $S^2$ is negligible since the cut edges are only slightly longer than non-cut edges and
as a result, taking lengths into account in the bisection phase does not offer much advantage.
We also observe that the number of samples required by the weighted $S^2$ method is close to the size of the cut $|\partial C|$ in most of the datasets.
Table~\ref{tab:cut_samples} also shows that the adaptive methods $S^2$ and weighted $S^2$ are very efficient at recovering the entire cut exactly, compared to the non-adaptive cutoff maximization method.
In practice, it is not necessary to reconstruct the signal exactly and some reconstruction error is allowed.
Figure~\ref{fig:error_samples} plots the classification error against the number of sampled nodes. It shows that
the classification error of the cutoff maximization method decreases rapidly in the early stages of sampling when very few samples are observed. 
However, the decrease is slow in later stages of sampling. $S^2$ methods, on the other hand, are good at reducing the error in the later stages, but performs poorly with only a few samples. The figure also shows that the hybrid method performs as well as the better method in each region. 

\vspace{-0.75\baselineskip}
\section{Conclusions}
\vspace{-0.25\baselineskip}
The paper generalizes the $S^2$ algorithm for the case of weighted graphs. The sampling complexity of the generalized algorithm is analyzed, and the gain over the unweighted version is demonstrated by simulation.
Additional experiments identify the region of tolerable reconstruction error in which the $S^2$ algorithms outperforms a graph frequency based global approach. A hybrid approach is proposed with the advantages of both methods. It remains open to analytically characterize of the gain of the weighted $S^2$ method over the unweighted version. Another interesting avenue for future work is to provide a performance analysis for the spectral sampling method which can suggest an optimal switching criterion for the hybrid method.
%


 
\newpage
\bibliographystyle{IEEEbib}
\bibliography{local,refs}

\end{document}